\documentclass[10pt,twocolumn,letterpaper]{article}

\usepackage{cvpr}
\usepackage{times}
\usepackage{epsfig}
\usepackage{graphicx}
\usepackage{amsmath}
\usepackage{amssymb}
\usepackage{multirow}
\usepackage{ctable}
\usepackage{caption}

\usepackage[breaklinks=true,bookmarks=false,colorlinks=true]{hyperref}

\cvprfinalcopy 

\def\cvprPaperID{0} 

\begin{document}
\title{BlendedMVS: A Large-scale Dataset for Generalized Multi-view Stereo Networks}

\author{
Yao Yao$^1$
\and
Zixin Luo$^1$
\and
Shiwei Li$^2$
\and
Jingyang Zhang$^1$
\and
Yufan Ren$^3$
\and
Lei Zhou$^1$
\and
Tian Fang$^2$
\and
Long Quan$^1$
\and \\
\centerline{$^1$The Hong Kong University of Science and Technology} 
\vspace{-1mm} \\
\centerline{\tt\small $\{$yyaoag, zluoag, jzhangbs, lzhouai, quan$\}$@cse.ust.hk} 
\vspace{2mm}\\
\centerline{$^2$Everest Innovation Technology \hspace{10mm} $^3$Zhejiang University}
\vspace{-1mm} \\
\centerline{\tt\small $\{$sli, fangtian$\}$@altizure.com \hspace{10mm} \tt\small renyufan@zju.edu.cn}
}

\maketitle

\begin{abstract}

While deep learning has recently achieved great success on multi-view stereo (MVS), limited training data makes the trained model hard to be generalized to unseen scenarios. Compared with other computer vision tasks, it is rather difficult to collect a large-scale MVS dataset as it requires expensive active scanners and labor-intensive process to obtain ground truth 3D structures. In this paper, we introduce BlendedMVS, a novel large-scale dataset, to provide sufficient training ground truth for learning-based MVS. To create the dataset, we apply a 3D reconstruction pipeline to recover high-quality textured meshes from images of well-selected scenes. Then, we render these mesh models to color images and depth maps. To introduce the ambient lighting information during training, the rendered color images are further blended with the input images to generate the training input. Our dataset contains over 17k high-resolution images covering a variety of scenes, including cities, architectures, sculptures and small objects. Extensive experiments demonstrate that BlendedMVS endows the trained model with significantly better generalization ability compared with other MVS datasets. The dataset and pretrained models are available at \url{https://github.com/YoYo000/BlendedMVS}.
\vspace{-2mm}

\end{abstract}

\section{Introduction}

Multi-view stereo (MVS) reconstructs the dense representation of the scene from multi-view images and corresponding camera parameters. While the problem is previously addressed by classical methods, recent studies~\cite{yao2018mvsnet,yao2019recurrent,ji2017surfacenet} show that learning-based approaches are also able to produce results comparable to or even better than classical state-of-the-arts. Conceptually, learning-based approaches implicitly take into account global semantics such as specularity, reflection and lighting information during the reconstruction, which would be beneficial for reconstructions of textureless and non-Lambertian areas. It has been reported on the small object DTU dataset~\cite{aanaes2016large} that, the best overall quality has been largely improved by recent learning-based approaches~\cite{yao2018mvsnet,yao2019recurrent,chenrui2019point,luoke2019pmvsnet}. 

By contrast, leaderboards of Tanks and Temples~\cite{knapitsch2017tanks} and ETH3D~\cite{schoeps2017cvpr} benchmarks are still dominated by classical MVS methods. In fact, current learning-based methods are all trained on DTU dataset~\cite{aanaes2016large}, which consists of small objects captured with a fixed camera trajectory. As a result, the trained model cannot generalize very well on other scenes. Moreover, previous MVS benchmarks~\cite{seitz2006comparison,strecha2008benchmarking,aanaes2016large,knapitsch2017tanks,schoeps2017cvpr} mainly focus on the point cloud evaluation rather than the network training. Compared with other computer vision tasks (e.g., classification and stereo), the training data for MVS reconstruction is rather limited, and it is desired to establish a new dataset to provide sufficient training ground truth for learning-based MVS.

\begin{figure*}
  \centering
  \vspace{-2mm}
  \includegraphics[angle=90,width=0.97\linewidth]{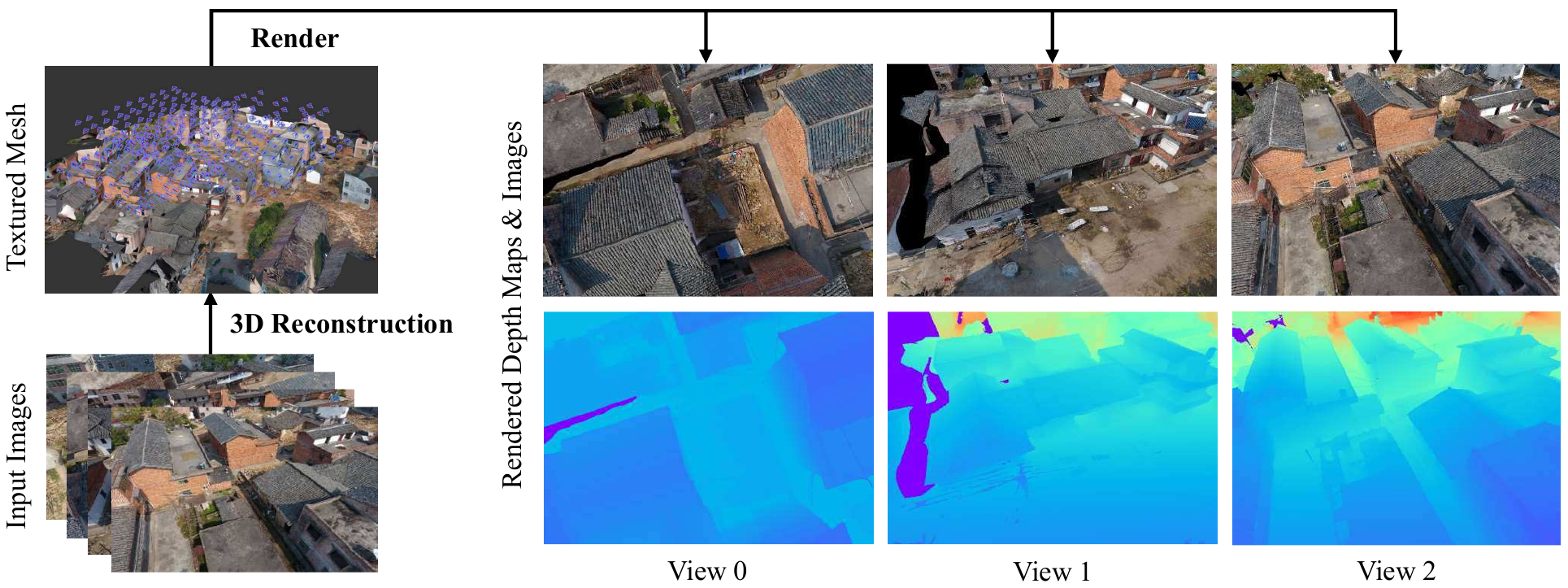}
  \vspace{-2mm}
  \caption{Pipeline of rendered data generation. We reconstruct the textured 3D model from input images, and then rendered the model into different view point to generate rendered images and depth maps.}
  \label{fig:render}
\end{figure*}

In this paper, we introduce BlendedMVS, a large-scale synthetic dataset for multi-view stereo training. Instead of using expensive active scanners to obtain ground truth point clouds, we propose to generate training images and depth maps by rendering textured 3D models to different viewpoints. The texture mesh of each scene is first reconstructed from images, which is then rendered into color images and depth maps. To introduce the ambient lighting information during training, we further blend rendered images with input color images to generate the training input. The resulting images inherit detailed visual cues from rendered color images, which makes them consistently align with rendered depth maps. At the same time, the blended images still largely preserve the realistic ambient lighting information from input images, which helps the trained model better generalize to real-world scenarios.

Our dataset contains 113 well selected and reconstructed 3D models. These textured models cover a variety of different scenes, including cities, architectures, sculptures and small objects. Each of the scene contains 20 to 1,000 input images and there are more than 17,000 images in total. We train recent MVSNet~\cite{yao2018mvsnet}, R-MVSNet~\cite{yao2019recurrent} and Point-MVSNet~\cite{chenrui2019point} on several MVS datasets. Extensive experiments on different validation sets demonstrate that models trained on BlendedMVS achieve better generalization ability compared with models trained on other MVS datasets. 

Our main contributions can be summarized as:
\begin{itemize}
\item We propose a low-cost data generation pipeline with a novel fusion approach to automatically generate training ground truth for learning-based MVS.

\item We establish the large-scale BlendedMVS dataset. All models in the dataset are well selected and cover a variety of diversified reconstruction scenarios.

\item We report on several benchmarks that BlendedMVS endows the trained model with significantly better generalization ability compared with other MVS datasets.
\end{itemize}

\section{Related Works}

\subsection{Learning-based MVS} 

Learning-based approaches for MVS reconstruction have recently shown great potentials. Learned multi-patch similarity~\cite{hartmann2017learned} first applies deep neural networks for MVS cost metrics learning. SurfaceNet~\cite{ji2017surfacenet} and DeepMVS~\cite{huang2018deepmvs} unproject images to the 3D voxel space, and use 3D CNNs to classify if a voxel belongs to the object surface. LSM~\cite{kar2017learning} and RayNet~\cite{paschalidou2018raynet} encoded the camera projection to the network, and utilized 3D CNNs or Markov Random Field to predict surface label. To overcome the precision deficiency in volume presentation, MVSNet~\cite{yao2018mvsnet} applies differentiable homography to build the cost volume upon the camera frustum. The network applies 3D CNNs for the cost volume regularization and regress the per-view depth map as output. The follow-up R-MVSNet~\cite{yao2019recurrent} is designed for high-resolution MVS, by replacing the memory-consuming the 3D CNNs with the recurrent regularization, and significantly reduce the peak memory size. More recently, Point-MVSNet~\cite{chenrui2019point} presents an point-based depth map refinement network, while MVS-CRF~\cite{xue2019mvscrf} introduces the conditional random field for the depth map refinement. 

\subsection{MVS Datasets}

Middlebury MVS~\cite{seitz2006comparison} is the earliest MVS dataset for MVS evaluation. It contains two indoor objects with low-resolution (640 $\times$ 480) images and calibrated cameras. Later, the EPFL benchmark~\cite{strecha2008benchmarking} captures ground truth models of building facades and provides high-resolution images (6.2 MP) and ground truth point clouds for MVS evaluation. To evaluate algorithms under different lighting conditions, DTU dataset~\cite{aanaes2016large} captures images and point clouds for more than 100 indoor objects with a fixed camera trajectory. The point clouds are further triangulated into mesh models and rendered into different view point to generate ground truth depth maps~\cite{yao2018mvsnet}. Current learning-based MVS networks~\cite{yao2018mvsnet, yao2019recurrent, chenrui2019point, luoke2019pmvsnet} usually apply DTU dataset as their training data. Recent Tanks and Temples benchmark~\cite{knapitsch2017tanks} captures indoor and outdoor scenes using high-speed video cameras, however, their training set only contains 7 scenes with ground truth point clouds. ETH3D benchmark~\cite{schoeps2017cvpr} contains one low-resolution set and one high-resolution set. But similar to Tanks and Temples, ETH3D only provides a small number of ground truth scans for the network training. The available training data in these datasets is rather limited, and a larger scale dataset is required to further exploit the potentials of learning-based MVS. In contrast, the proposed dataset will provide more than 17,000 images with ground truth depth maps, which covers a variety of diversified scenes and can greatly improve the generalization ability of the trained model.

\subsection{Synthetic Datasets}

Generating synthetic datasets for training is a common practice in many computer vision tasks, as a large amount of ground truth can be generated at very low cost. Thanks to recent advances in computer graphics, the rendering effect becomes increasingly photo-realistic, making the usage of synthetic datasets more plausible. For example, synthetic rendered images are used in stereo matching~\cite{butler2012sintel, mayer2016large, zhang2018unreal}, optical flow~\cite{butler2012sintel, mayer2016large, gaidon2016virtual}, object detection~\cite{handa2016understanding, tremblay2018training} and semantic segmentation~\cite{handa2016understanding, richter2016playing, gaidon2016virtual, ros2016synthia, straub2019replica}. Similar to these datasets, we consider incorporating the lighting effects in rendering synthetic datasets for 3D reconstruction. However, since it is difficult to generate correct material properties in different parts of the model, we resort to a blending approach with original images to recover the lighting effects.

\section{Dataset Generation}

The proposed data generation pipeline is shown in Fig.~\ref{fig:render}. We first apply a full 3D reconstruction pipeline to produce the 3D textured mesh from input images (Sec.~\ref{sec:pipeline}). Next, the mesh is rendered to each camera view point to obtain the rendered image and the corresponding depth map. The final training image input is generated by blending the rendered image and input image in our proposed manner (Sec.~\ref{sec:blending}).

\subsection{Textured Mesh Generation} \label{sec:pipeline}

The first step to build a synthetic MVS dataset is generating sufficient high-quality textured mesh models. Given input images, we use Altizure online platform~\cite{Altizure} for the textured mesh reconstruction. The software will perform the full 3D reconstruction pipeline and return the textured mesh and camera poses as final output.

With the textured mesh model and camera positions of all input images, we then render the mesh model to each camera view point to generate the rendered images and rendered depth maps. One example is shown in Fig.~\ref{fig:render}. The rendered depth maps will be used as the ground truth depth maps during training.

\begin{figure}
  \centering
  \includegraphics[width=0.92\linewidth]{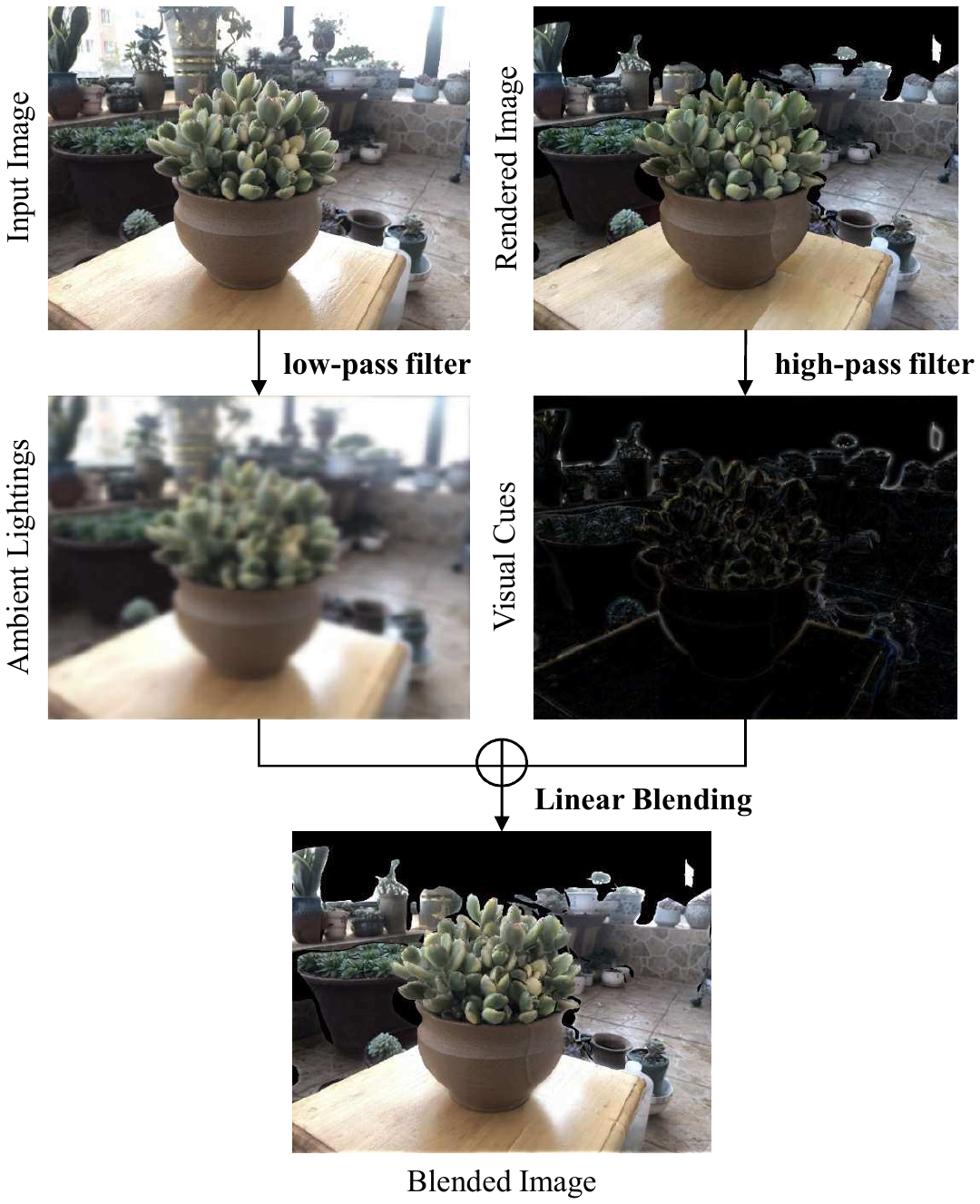}
  \caption{The blending process of the data generation pipeline. The high-pass filter is applied to extract image visual cues from the rendered image, while the low-pass filter is applied to extract ambient lightings from the input.}
  \label{fig:blending}
  \vspace{-3mm}
\end{figure}

\subsection{Blended Image Generation} \label{sec:blending}

Intuitively, rendered images and depth maps can be directly used for the network training. However, one potential problem is that rendered images do not contain view-dependent lightings. In fact, a desired training sample to multi-view stereo network should satisfy:
\begin{itemize}
\item Images and depth maps should be consistently aligned. The training sample should provide reliable mappings from input images to ground truth depth maps.
\item Images should reflect view-dependent lightings. The realistic ambient lighting could strengthen model's generalization ability to real-world scenarios.
\end{itemize}
To introduce lightings to rendered images, one solution is to manually assign mesh materials and set up lighting sources during the rendering process. However, this is extremely labor-intensive, which makes it rather difficult to build a large-scale dataset. 

On the other hand, the original input images have already contained the natural lighting information. The lighting could be automatically overlaid to rendered images if we can directly extract such information from input images. Specifically, we notice that ambient lightings are mostly \textbf{low-frequency} signals in images, while visual cues for establishing multi-view dense correspondences (e.g., rich textures) are mostly \textbf{high-frequency} signals in images. Following the observation, we propose to extract visual cues from the rendered image $\mathbf{I}_{r}$ using a high-pass filter $\mathbf{H}$, and extract the view-dependent lighting from the input image $\mathbf{I}$ using the low-pass filters $\mathbf{L}$. The visual cues and lightings are fused to generate the blended image $\mathbf{I}_{b}$ (Fig.~\ref{fig:blending}):
\begin{equation}
\begin{split}
\mathbf{I}_{b} &= \mathbf{I}_{r} * \mathbf{H} + \mathbf{I} * \mathbf{L} \\
&= \mathcal{F}^{-1} \big( \mathcal{F}(\mathbf{I}_{r}) \cdot \mathbf{H}_{f} \big) + \mathcal{F}^{-1} \big( \mathcal{F}(\mathbf{I}) \cdot \mathbf{L}_{f} \big)
\end{split}
\end{equation}
where `$*$' denotes the convolution operation, `$\cdot$' the element-wise multiplication. The symbols $\mathcal{F}$ and $\mathcal{F}^{-1}$ are 2D Fast Fourier Transformation (FFT) and inverse FFT respectively. In our implementation, the filtering process is performed in the frequency domain. $\mathbf{L}_f$ and $\mathbf{H}_f$ are approached by 2D Gaussian low-pass and high-pass filters:
\begin{equation}
\mathbf{L}_f(u, v) = \exp{\big(- \frac{(u^2 + v^2)}{2 \cdot D_0}\big)}
\end{equation}
\begin{equation}
\mathbf{H}_f(u, v) = 1 - \mathbf{L}_f(u, v)
\end{equation}

\begin{figure}
  \centering
  \includegraphics[width=1.0\linewidth]{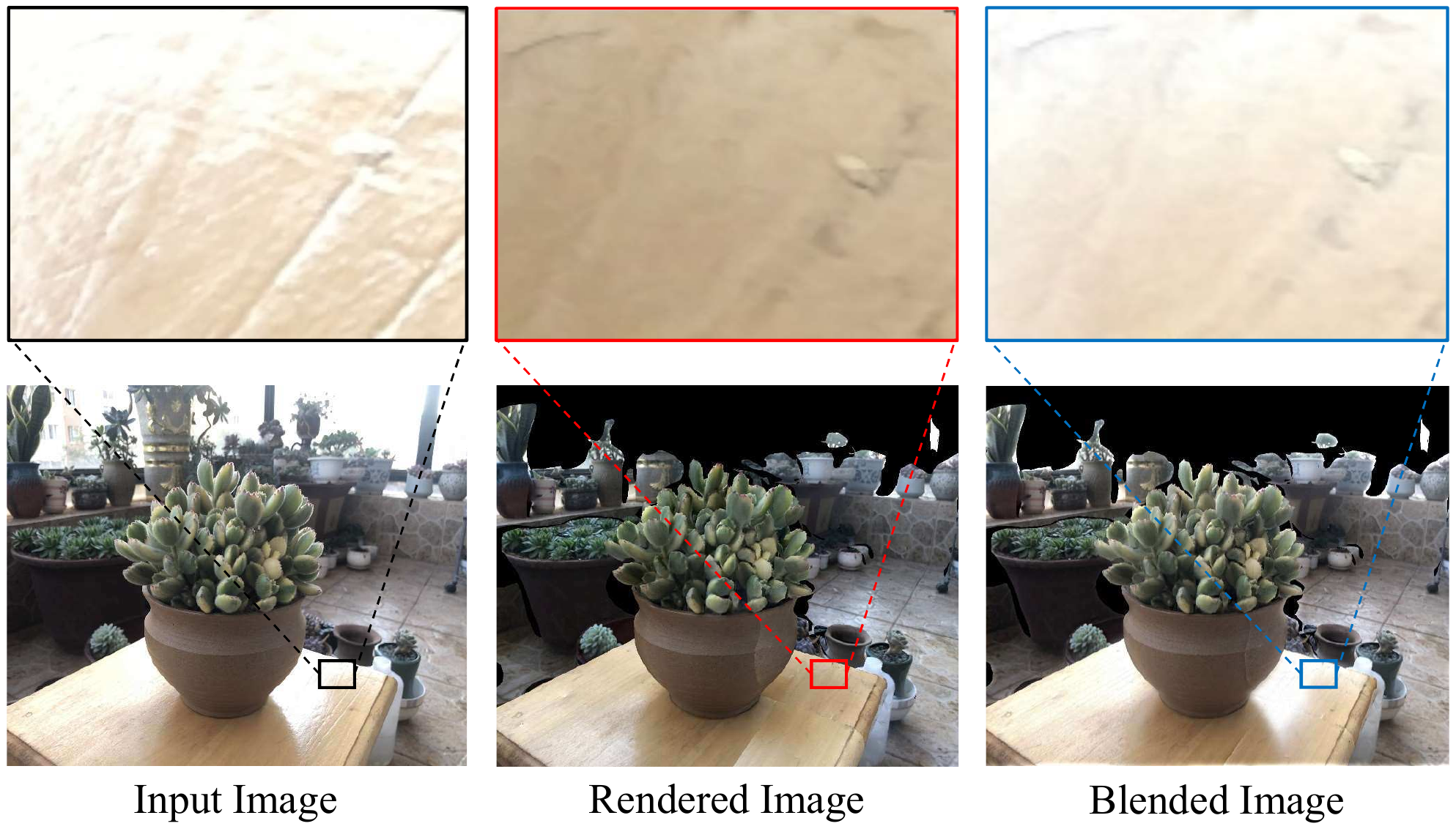}
  \caption{Detailed textures of input, rendered and blended images. The blended image has similar background lightings to the input image, while inherits texture details from the rendered image.}
  \label{fig:textures}
\end{figure}

The Gaussian kernel factor is empirically set to $D_0 = 5,000$ in our experiments. The blended image inherits detailed visual cues from the rendered image, while at the same time largely preserves realistic environmental lightings from the input image. Fig.~\ref{fig:textures} illustrates the differences between these three images. We will demonstrate in Sec.~\ref{sec:ablation} that models trained with blended images have better generalization abilities to different scenes.

\begin{figure*}
  \centering
  \includegraphics[width=1.0\linewidth]{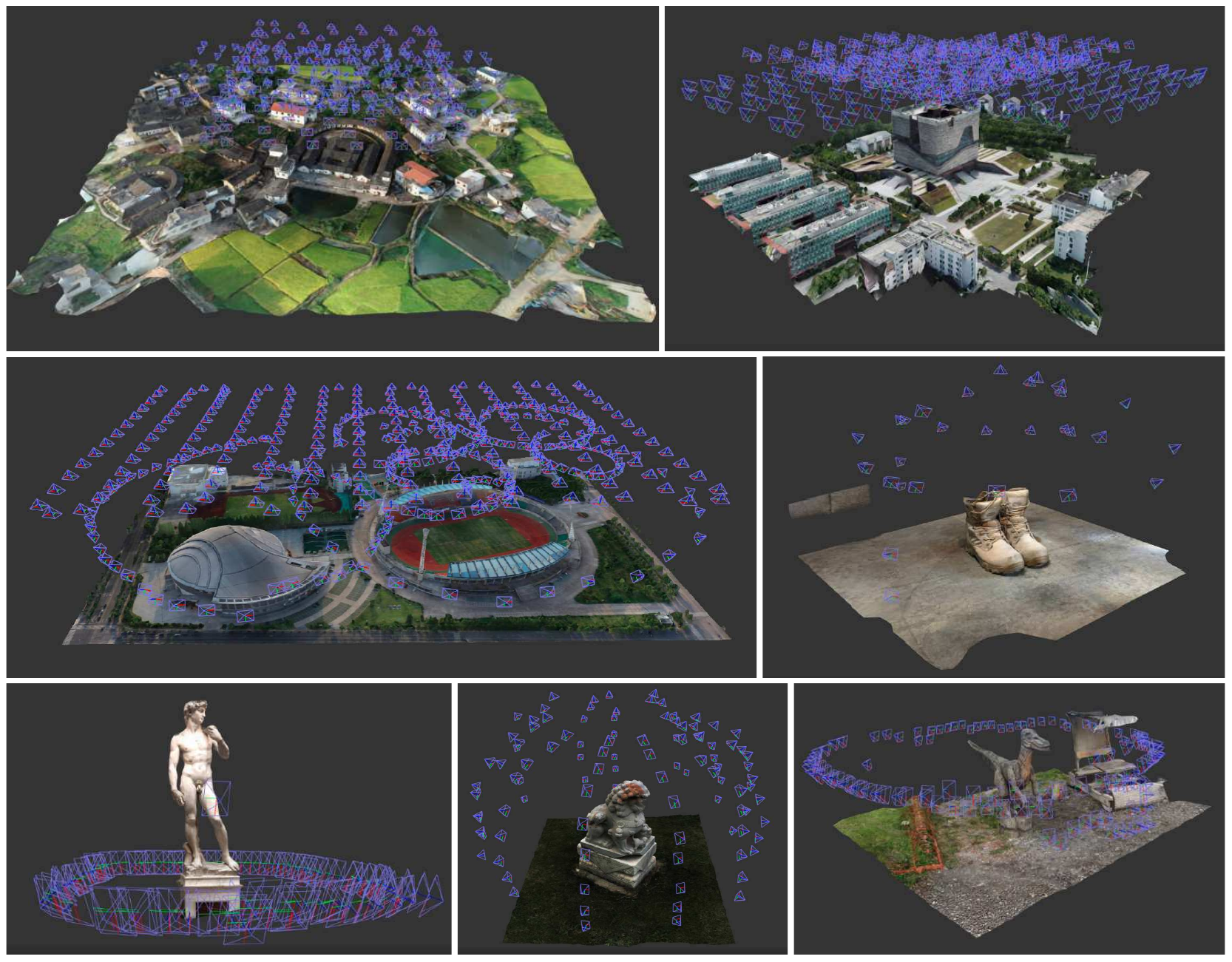}
  \vspace{-3mm}
  \caption{Several textured models with camera trajectories in BlendedMVS dataset. The blue box indicate the camera position in the 3D space. Our dataset contains 113 scenes in total.}
  \label{fig:scenes}
  \vspace{-3mm}
\end{figure*}

\section{Scenes and Networks}

\subsection{Scenes} \label{sec:scenes}

For the content of the proposed dataset, we manually select 113 well-reconstructed models publicly available in the Altizure.com online platform. These models cover a variety of different scenes, including architectures, street-views, sculptures and small objects. Each of the scene contains 20 to 1,000 input images, and totally there are 17,818 images in the whole dataset. It is also noteworthy that unlike DTU dataset~\cite{aanaes2016large} where all scenes are captured by a fixed robot arm, scenes in BlendedMVS contain a variety of different camera trajectories. The unstructured camera trajectories can better model different image capturing styles, and is able to make the network more generalizable to real-world reconstructions. Fig.~\ref{fig:scenes} shows 7 scenes in BlendedMVS dataset with camera positions.

The dataset also provides training images and ground truth depth maps with a unified image resolution of $H \times W = 1536 \times 2048$. As input images are usually with different resolutions, we first resize all blended images and rendered depth maps to a minimum image size $H_s \times W_s$ such that $H_s >= 1536$ and $W_s >= 2048$. Then, we crop image patches of size $H \times W = 1536 \times 2048$ from the resized image centers to build training samples for BlendedMVS dataset. The corresponding camera parameters are changed accordingly. Also, the depth range is provided for each image as this information is usually required by depth map estimation algorithms.

\vspace{-2mm}
\paragraph{Online Augmentation}
We also augment the training data during the training process. The following photometric augmentations are considered in our training: 1) Random brightness: we change the brightness of each image by adding a random value $b$ such that $-50 < b < 50$, and then clip the image intensity value to the standard range of $0-255$. 2) Random contrast: we change the contrast of each image with a random contrast factor $c$ such that $0.3 < c < 1.5$, and then clip the image to the standard range of $0-255$. 3) Random motion blur: we add the Gaussian motion blur to each input image. We consider a random motion direction and a random motion kernel size of $m = 1$ or $3$ during the augmentation. The above mentioned augmentations will be imposed to each training image in a random order. In the ablation study section~\ref{sec:ablation}, we will demonstrate the improvement brought by the online augmentation.

\subsection{Networks}

To verify the effectiveness of the proposed dataset, we train and evaluate recent MVSNet~\cite{yao2018mvsnet}, R-MVSNet~\cite{yao2019recurrent} and Point-MVSNet \cite{chenrui2019point} on BlendedMVS dataset.

\vspace{2mm}
\noindent\textbf{MVSNet}~\cite{yao2018mvsnet} is an end-to-end deep learning architecture for depth map estimation from multiple images. Given a reference image $\mathbf{I}_1$ and several source images $\{\mathbf{I}_i\}_{i=2}^N$, MVSNet first extract deep image features $\{\mathbf{F}_i\}_{i=1}^N$ for all images. Next, image features are warped into the reference camera frustum to build the 3D feature volumes $\{\mathbf{V}_i\}_{i=1}^N$ through the differentiable homographies. The network applies a variance-based cost metric to build the cost volume $\mathbf{C}$ and applies a multi-scale 3D CNNs for the cost volume regularization. The depth map $\mathbf{D}$ is regressed from the volume through the soft argmin \cite{kendall2017end} operation. The network is trained with the stander L1 loss function. 

\vspace{2mm}
\noindent\textbf{R-MVSNet}~\cite{yao2019recurrent} is an extended version of MVSNet for high-resolution MVS reconstruction. Instead of regularizing the whole 3D cost volume $\mathbf{C}$ with 3D CNNs at once, R-MVSNet applies the recurrent neural network to sequentially regularize the 2D cost maps $\mathbf{C}(d)$ through the depth direction, which dramatically reduces the memory consumption for MVS reconstruction. Meanwhile, R-MVSNet treats depth map estimation as a classification problem and applies the cross-entropy loss during the network training.

\vspace{2mm}
\noindent\textbf{Point-MVSNet}~\cite{chenrui2019point} is a point-based deep framework for MVS reconstruction. In the network, the authors apply MVSNet framework to generate a coarse depth map, convert it into a point cloud and finally refine the point cloud iteratively by estimating the residual between the depth of the current iteration and that of the ground truth. 

\vspace{2mm}
\noindent\textbf{Implementations} we directly use the open-source implementations of the three networks from their GitHub pages. Compared with the original papers, several modifications have been made to MVSNet and R-MVSNet: 1) The 5-layer 2D CNNs is replaced by a 2D U-Net to enlarge the receptive field during image feature extraction. 2) The batch normalization \cite{ioffe2015batch} is replaced with the group normalization \cite{wu2018group} with fixed a group channel size of 8 to improve the network performance when training with small batch size. 3) The refinement network in MVSNet is removed as this part only brings limited performance gain. 4) The variational refinement step in R-MVSNet is removed so as to avoid the non-learning component affecting the dataset evaluation.

All models are trained using one GTX 2080 Ti GPU with $batchsize = 1$. MVSNet and R-MVSNet are trained for 160k iterations, while Point-MVSNet is trained for 320k iterations with the first 100k for coarse MVSNet initialization and another 220k for the end-to-end training.

\section{Experiments}

\subsection{Quantitative Evaluation} \label{sec:comparison}

\subsubsection{Depth Map Validation}

To demonstrate the capacity of BlendedMVS dataset, we compare models trained on 1) DTU training set, 2) ETH3D low-res training set, 3) MegaDepth dataset and 4) BlendedMVS training set. Evaluations are done on the corresponding validation sets. Three metrics are considered in our experiments: 1) the end point error (EPE), which is the average $L_1$ loss between the inferred depth map and the ground truth depth map; 2) the $>1$ pixel error, which is the ratio of pixels with $L_1$ error larger than 1 depth-wise pixel; and 3)  the $>3$ pixel error. Quantitative results are shown in Fig.~\ref{fig:validation}.

\vspace{3mm}
\noindent \textbf{Trained on DTU \cite{aanaes2016large}} 
As suggested by previous methods \cite{yao2018mvsnet, ji2017surfacenet, yao2019recurrent}, DTU dataset is divided into training, validation and evaluation sets. We train the three networks with a fixed input sample size of $H \times W \times D = 512 \times 640 \times 128$ and fixed depth range of $[d_{min}, d_{max}] = [425, 937]$. 

It is reported in Fig.~\ref{fig:validation} that all three models trained on DTU (black lines) perform very well on DTU validation set, however, produce high validation errors in BlendedMVS and ETH3D datasets. In fact, models are overfitted in small-scale indoor scenes, showing the importance of having rich object categories in MVS training data.

\vspace{3mm}
\noindent \textbf{Trained on ETH3D \cite{schoeps2017cvpr}} 
The ETH3D training set contains 5 scenes. To separate the training and the validation, we take \textit{delivery\_area}, \textit{electro}, \textit{forest} as our training scenes, and \textit{playground}, \textit{terrains} as our validation scenes. The training sample size is fixed to $H \times W \times D = 480 \times 896 \times 128$. The per-view depth range is determined by the sparse point cloud provided by the dataset. 

As shown in Fig.~\ref{fig:validation}, validation errors of models trained on ETH3D (blue dash lines) are high in all validation sets including its own dataset, indicating that ETH3D training set does not provide sufficient data for MVS training.

\vspace{3mm}
\noindent \textbf{Trained on MegaDepth \cite{li2018megadepth}} 
MegaDepth dataset is originally built for single-view depth map estimation that it applies multi-view depth map estimation to generate the depth training data. The dataset provides image-depthmap training pairs and SfM output files from COLMAP \cite{schonberger2016structure}. To apply MegaDepth for the MVS training, we apply the view seletion and the depth range estimation \cite{yao2018mvsnet,yao2019recurrent} to generate training files in MVSNet format. Also, as reconstructed depth maps of crowdsourced images are usually incomplete, we only use those training samples with more than $20\%$ valid pixels in the reference depth map during our training. There are 39k MVS training samples in MegaDepth dataset after the proposed pre-processing. The training input size is fixed to $H \times W \times D = 512 \times 640 \times 128$ by applying the resize-and-crop strategy as described in~\ref{sec:scenes}. 

Although MegaDepth contains more training samples than BlendeMVS, models trained on MegaDepth (green dash lines in Fig.~\ref{fig:validation}) are still inferior to models trained on BlendedMVS. We believe there are two major problems of applying MegaDepth for the MVS training: 1) the ground truth depth map is generated through MVS reconstructions. In this case, input images and reconstructed depth maps are not consistently aligned and the network will tend to overfit to the chosen algorithm \cite{schonberger2016pixelwise}. 2) MegaDepth is built upon crowdsourced internet photos. The crowdsourced images are not well-captured and the training data quality could have significant influences on the training result.

\vspace{3mm}
\noindent \textbf{Trained on BlendedMVS} 
To train MVS networks with BlendedMVS, we resize all training samples to $H \times W = 576 \times 768$ for MVSNet and R-MVSNet, and further crop the samples to $H \times W = 448 \times 768$ for Point-MVSNet. The depth sample number is set to $D = 128$. Our dataset is also divided into 106 training scenes and 7 validation scenes to evaluate the network training. 

As shown in Fig.~\ref{fig:validation}, models trained on BlendedMVS (red lines) generalizes well to both DTU and ETH3D scenes. All models achieve the best validation results on BlendedMVS and ETH3D validation sets, and achieve the second best result (very close to the best) on DTU validation set, showing the strong generalization ability brought by our dataset.

\begin{figure*}
  \begin{minipage}{0.97\linewidth}
  \centering
  \includegraphics[width=1\linewidth]{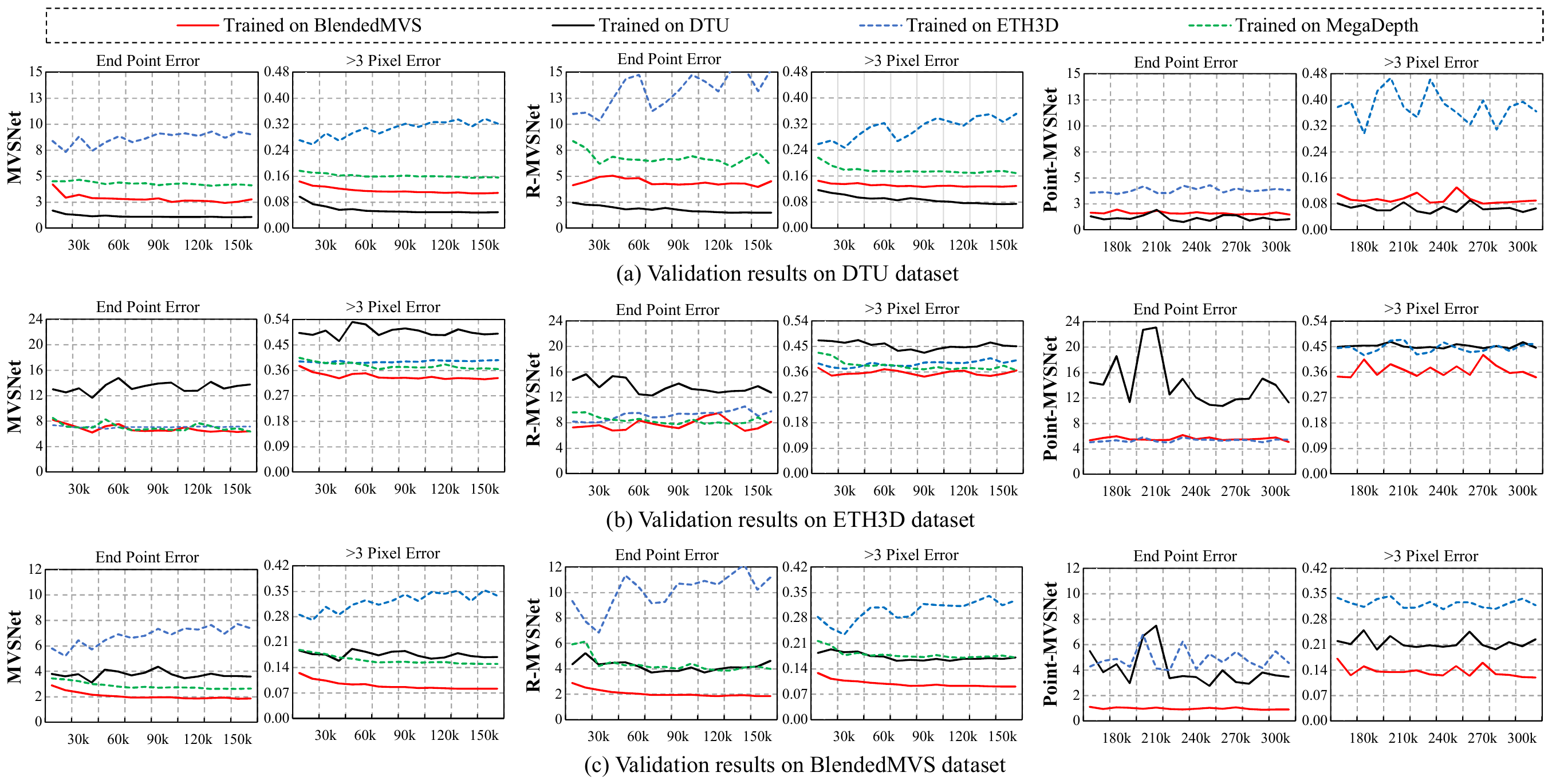}
  \vspace{-7mm}
  \caption{\textbf{Depth map validation errors} during the training process on all validation sets. Results of models trained on BlendedMVS (red lines) demonstrate good generalization ability on both DTU and ETH3D validation sets.}
  \label{fig:validation}
  \end{minipage}
  
  \vspace{5mm}
  \begin{minipage}{1\linewidth}
\resizebox{0.90\textwidth}{!}{%
\begin{tabular}{c|c|ccccccc|c}
\specialrule{.2em}{.1em}{.1em}
R-MVSNet Models	            		& Metrics   		& Barn 	& Caterpillar & Church& Courthouse& Ignatius& Meetingroom& Truck & Average 	\\ \hline
\multirow{3}{*}{Trained on DTU~\cite{aanaes2016large}} & \textit{Precision}& 0.387	& 0.301	& 0.498 & 0.399 & 0.409 & 0.391 & 0.559 & 0.421	\\ 
                            			& \textit{Recall}	& 0.674	& 0.755	& 0.313 & 0.731 & 0.856 & 0.213 & 0.846 & 0.623	\\  
                            			& \textit{F\_score} & 0.492	& 0.430	& 0.384 & 0.517 & 0.553 & 0.276 & 0.673 & 0.475 \\ \hline
\multirow{3}{*}{Trained on ETH3D~\cite{schoeps2017cvpr}}  	& \textit{Precision}& 0.334 & 0.297 & 0.497 & 0.347 & 0.362 & 0.324 & 0.492 & 0.379 \\ 
                            			& \textit{Recall} 	& 0.564 & 0.608 & 0.221 & 0.598 & 0.750 & 0.112 & 0.706 & 0.508 \\
                            			& \textit{F\_score} & 0.420 & 0.399 & 0.306 & 0.439 & 0.488 & 0.166 & 0.580 & 0.400 \\ \hline
\multirow{3}{*}{Trained on MegaDepth~\cite{li2018megadepth}} 	& \textit{Precision}& 0.414	& 0.291	& 0.566 	& 0.441 	& 0.408 	& 0.418 	& 0.522 	& 0.437	\\ 
                            			& \textit{Recall}	& 0.676	& 0.724	& 0.282 	& 0.741 	& 0.854 	& 0.152 	& 0.815 	& 0.606	\\  
                            			& \textit{F\_score} & 0.513	& 0.415	& 0.376 	& 0.553 	& 0.552 	& 0.223 	& 0.636 	& 0.467 \\ \hline
\multirow{3}{*}{Trained on BlendedMVS} 	& \textit{Precision}& 0.432	& 0.352	& 0.570 	& 0.462	& 0.492 & 0.444 & 0.602 & 0.479	\\ 
                            			& \textit{Recall}	& 0.715	& 0.770	& 0.387 	& 0.765 	& 0.901 & 0.251 & 0.845 & 0.662	\\  
                            			& \textit{F\_score} & \textbf{0.539}	& \textbf{0.484}	& \textbf{0.461} & \textbf{0.577} & \textbf{0.636} & \textbf{0.321} & \textbf{0.703} & \textbf{0.532} \\ 
\specialrule{.2em}{.1em}{.1em}
\end{tabular}%
}
\vspace{-1mm}
\centering
\captionof{table}{\textbf{Point cloud evaluations} on \textit{Tanks and Temples} training set~\cite{knapitsch2017tanks}. R-MVSNet trained on BlendedMVS outperforms models trained on other datasets in all scenes.}
\label{tab:pointcloud}
  \end{minipage}
  
  \vspace{5mm}
  \begin{minipage}{1\linewidth}
  \centering
  \includegraphics[width=0.96\linewidth]{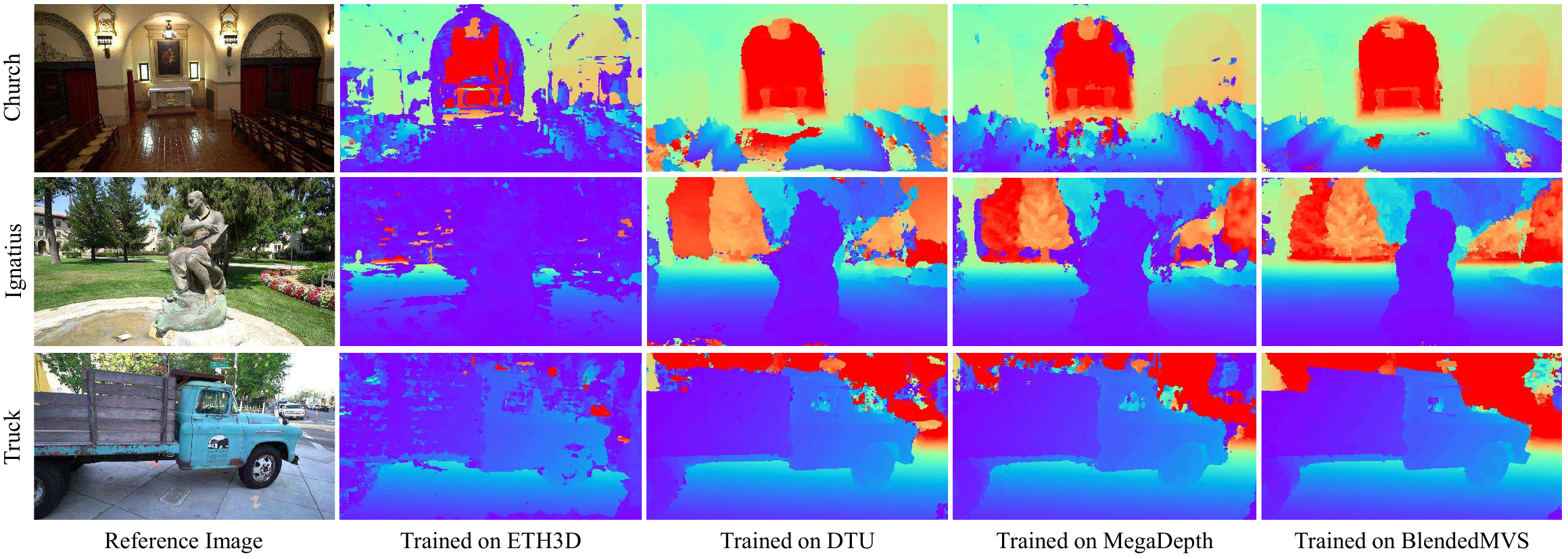}
  \vspace{-3mm}
  \caption{Qualitative comparisons on depth map reconstructions using R-MVSNet~\cite{yao2019recurrent}. The model trained on BlendedMVS generates much cleaner results than models trained on the other three datasets.}
  \label{fig:depthmap}
  \end{minipage}
\end{figure*}

\subsubsection{Point Cloud Evaluation}

We also compare point cloud reconstructions of models trained on DTU, ETH3D, MegaDepth and BlendedMVS on Tanks and Temples~\cite{knapitsch2017tanks} training set. As the dataset contains wide-depth-range scenes that cannot be handled by MVSNet and PointMVSNet, we only test R-MVSNet (trained for 150k iterations) in this experiment. We follow methods described in R-MVSNet paper to recover camera parameters of input images, and then perform the per-view source image selection and depth range estimation based on the sparse point cloud. For post-processing, we also follow previous works~\cite{yao2018mvsnet,yao2019recurrent} to apply the visibility-based depth map fusion~\cite{merrell2007real}, average depth map fusion and visibility depth map filter to generate the 3D point cloud.

The dataset reports three evaluation metrics, namely \textit{precision} (\textit{accuracy}), \textit{recall} (\textit{completeness}) and the overall \textit{f\_score}~\cite{knapitsch2017tanks, schoeps2017cvpr} to quantitatively measure the reconstruction quality. As shown in Table~\ref{tab:pointcloud}, R-MVSNet trained on DTU~\cite{aanaes2016large} and MegaDepth~\cite{li2018megadepth} achieve similar \textit{f\_score} performances, while R-MVSNet trained on the proposed dataset outperforms models trained on the other three datasets for all scenes. The average \textit{f\_score} is improved from $0.475$ to $0.532$ by simply replacing the training data from DTU to BlendedMVS. Qualitative comparisons on depth maps are shown in Fig.~\ref{fig:depthmap}.

\begin{table}[]
\resizebox{0.48\textwidth}{!}{%
\begin{tabular}{c|r|ccc}
\specialrule{.2em}{.1em}{.1em}
\multicolumn{1}{c|}{Networks} &  Training Images & \multicolumn{1}{l}{EPE} & \multicolumn{1}{l}{\textless{}1 Px. Err} & \multicolumn{1}{l}{\textless{}3 Px. Err} \\ \hline
\multirow{6}{*}{MVSNet~\cite{yao2018mvsnet}}   	& Rendered   	& 2.99     		& 0.245			& 0.136			\\ 
                          						& Input       	& 3.70      		& 0.243			& 0.135         	\\ 
                          						& Blended       	& 2.88     	 	& 0.224			& 0.118			\\ \cline{2-5} 
                          						& Rendered+Aug.	& 2.94      		& 0.225			& 0.116			\\ 
                          						& Input+Aug.    	& 3.16    		& 0.234			& 0.123			\\ 
                          						& Blended+Aug.  	& \textbf{2.53}	& \textbf{0.219}	& \textbf{0.107}	\\ \hline
\multirow{6}{*}{R-MVSNet~\cite{yao2019recurrent}}& Rendered      	& 5.54          	& 0.251			& 0.148			\\ 
                          						& Input         	& 4.47          	& 0.242			& 0.134     	  	\\
                         						& Blended     	& 5.77    		& 0.239			& 0.137			\\ \cline{2-5} 
                          						& Rendered+Aug. 	& 5.10			& 0.	238			& 0.132			\\ 
                          						& Input+Aug.    	& \textbf{3.86}	& 0.241			& \textbf{0.126}	\\ 
                          						& Blended+Aug. 	& 3.95    		& \textbf{0.234}	& \textbf{0.127}	\\ 
\specialrule{.2em}{.1em}{.1em}
\end{tabular}%
}
  \vspace{-2mm}
\centering
\caption{Ablation study on using different images for training. Validation errors on DTU dataset~\cite{aanaes2016large} show that blended images with online augmentation produces the best result.}
\label{tab:ablation}
  \vspace{-2mm}
\end{table}

\subsection{Ablation Study on Training Image} \label{sec:ablation}

Next, we study the differences of using 1) input images, 2) rendered images and 3) blended images as our training images. For these three setting, we also study the effectiveness of the online photometric augmentation. All models are trained for 150k iterations and are validated on DTU validation set. Comparison results are shown in Table~\ref{tab:ablation}.

\vspace{1mm}
\noindent \textbf{Environmental Lightings} The proposed setting of blended images with photometric augmentation produces the best result, while rendered images only produces the worst result among all. Also, all images with photometric augmentation results in lower validation errors than without, showing that view-dependent lightings are indeed important for MVS network training. 

\vspace{1mm}
\noindent \textbf{Training with Input Images} It is noteworthy that while input images are not completely consistent with rendered depth maps, training R-MVSNet with input images (with or without the augmentation) also produces satisfying results (Table~\ref{tab:ablation}). The reason might be that 3D structures have been correctly recovered for most of the scenesc as all scenes are well-selected in advance. In this case, rendered depth maps can be regarded as the semi ground truth given input images, which could be jointly used for MVS network training.

\begin{figure}
  \centering
  \includegraphics[width=1.0\linewidth]{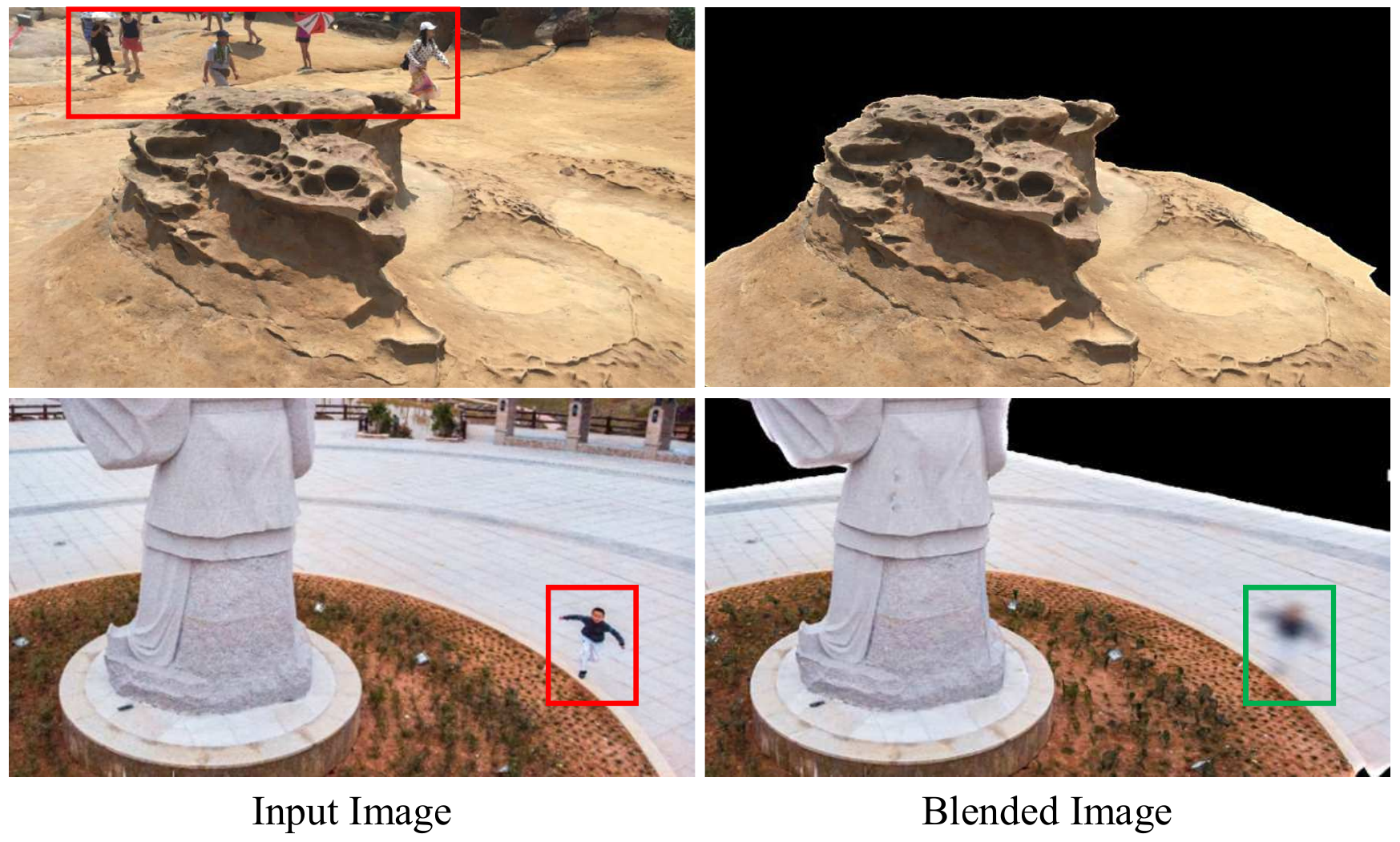}
  \vspace{-5mm}
  \caption{Privacy preserving with blended images. Humans will be removed or blurred in the blended images.}
  \label{fig:privacy}
\end{figure}

\subsection{Discussions} \label{discussion}

\noindent \textbf{Imperfect Reconstruction} One concern about using the reconstructed model for the MVS training is whether defects or imperfect reconstructions in textured models would affect the training process. In fact, blended images inherit detailed visual cues from rendered images, which are always consistent with rendered depth maps even if defects occur. In this case, the training process will not be deteriorated. 

For the same reason, we could change the Altizure online platform to any other 3D reconstruction pipelines to recover the mesh model. What we have presented is a low-cost MVS training data generation pipeline that does not rely on any particular textured model reconstruction method.
 
\vspace{1mm}
\noindent \textbf{Occlusion and Normal Information} While current learning-based approaches~\cite{yao2018mvsnet, yao2019recurrent, chenrui2019point, luoke2019pmvsnet} does not take into account the pixel-wise occlusion and normal information, our dataset provides such ground truth information as well. The occlusion and normal information could be useful for future visibility-aware and patch-based MVS networks.

\vspace{1mm}
\noindent \textbf{Privacy} Using blended images could also help preserve the data privacy. For example, pedestrians in input images are usually dynamic, which will not be reconstructed in the textured model and rendered images (first row in Fig.~\ref{fig:privacy}). Furthermore, if pedestrians appear in front of the reconstructed object, our image blending process will only extract blurred human shapes from the input image, which helps conceal user identities in the blended image (second row in Fig.~\ref{fig:privacy}). 

\subsection{Applications}
While BlendedMVS is originally designed for multi-view stereo network training, it is also applicable to a variety of 3D geometry related tasks, including, but not limited to 1) image retrieval \cite{shen2018mirror}, 2) Image feature detection and description \cite{luo2018geodesc,luo2019contextdesc,luo2020aslfeat}, 3) single-view/multi-view camera pose regression, 4) single-view depth/normal estimation and 5) disparity estimation. 

Based on BlendedMVS, we have progressively built BlendedMVG, a multi-purpose large-scale dataset, to solve general 3D geometry related problems. Except for the existing 113 scenes, we add 389 more scenes to the new dataset. BlendedMVG totally contains over 110k training samples, and has been successfully applied to image feature detection and description learning \cite{luo2020aslfeat}.

\section{Conclusion}

We have presented the BlendedMVS dataset for MVS network training. The proposed dataset provides more than 17k high-quality training samples covering a variety of scenes for multi-view depth estimation. To build the dataset, we have reconstructed textured meshes from input images, and have rendered these models into color images and depth maps. The rendered color image has been further blended with the input image to generate the training image input. We have trained recent MVS networks using BlendedMVS and other MVS datasets. Both quantitative and qualitative results have demonstrated that models trained on BlendedMVS achieve significant better generalization abilities than models trained on other datasets.

\section{Acknowledgments}

This work is supported by Hong Kong RGC GRF
16206819, Hong Kong RGC GRF
16203518 and Hong Kong T22-603/15N.
We thank Rui Chen for helping train and validate PointMVSNet~\cite{chenrui2019point} in our dataset.

\clearpage

{\small
\bibliographystyle{ieee_fullname}
\bibliography{egbib}

\begin{thebibliography}{10}\itemsep=-1pt

\bibitem{Altizure}
Altizure: Mapping the world in 3d.
\newblock \url{https://www.altizure.com}.

\bibitem{aanaes2016large}
Henrik Aan{\ae}s, Rasmus~Ramsb{\o}l Jensen, George Vogiatzis, Engin Tola, and
  Anders~Bjorholm Dahl.
\newblock Large-scale data for multiple-view stereopsis.
\newblock In {\em International Journal of Computer Vision (IJCV)}, 2016.

\bibitem{butler2012sintel}
Daniel~J. Butler, Jonas Wulff, Garrett~B. Stanley, and Michael~J. Black.
\newblock A naturalistic open source movie for optical flow evaluation.
\newblock In {\em European Conference on Computer Vision (ECCV)}, 2012.

\bibitem{chenrui2019point}
Rui Chen, Songfang Han, Jing Xu, and Hao Su.
\newblock Point-based multi-view stereo network.
\newblock In {\em International Conference on Computer Vision (ICCV)}, 2019.

\bibitem{gaidon2016virtual}
Adrien Gaidon, Qiao Wang, Yohann Cabon, and Eleonora Vig.
\newblock Virtual worlds as proxy for multi-object tracking analysis.
\newblock In {\em Computer Vision and Pattern Recognition}, 2016.

\bibitem{handa2016understanding}
Ankur Handa, Viorica Patraucean, Vijay Badrinarayanan, Simon Stent, and Roberto
  Cipolla.
\newblock Understanding real world indoor scenes with synthetic data.
\newblock In {\em Computer Vision and Pattern Recognition (CVPR)}, 2016.

\bibitem{hartmann2017learned}
Wilfried Hartmann, Silvano Galliani, Michal Havlena, Luc Van~Gool, and Konrad
  Schindler.
\newblock Learned multi-patch similarity.
\newblock In {\em International Conference on Computer Vision (ICCV)}, 2017.

\bibitem{huang2018deepmvs}
Po-Han Huang, Kevin Matzen, Johannes Kopf, Narendra Ahuja, and Jia-Bin Huang.
\newblock Deepmvs: Learning multi-view stereopsis.
\newblock In {\em Computer Vision and Pattern Recognition (CVPR)}, 2018.

\bibitem{ioffe2015batch}
Sergey Ioffe and Christian Szegedy.
\newblock Batch normalization: Accelerating deep network training by reducing
  internal covariate shift.
\newblock In {\em International Conference on Machine Learning (ICML)}, 2015.

\bibitem{ji2017surfacenet}
Mengqi Ji, Juergen Gall, Haitian Zheng, Yebin Liu, and Lu Fang.
\newblock Surfacenet: An end-to-end 3d neural network for multiview stereopsis.
\newblock In {\em International Conference on Computer Vision (ICCV)}, 2017.

\bibitem{kar2017learning}
Abhishek Kar, Christian H{\"a}ne, and Jitendra Malik.
\newblock Learning a multi-view stereo machine.
\newblock In {\em Advances in Neural Information Processing Systems (NIPS)},
  2017.

\bibitem{kendall2017end}
Alex Kendall, Hayk Martirosyan, Saumitro Dasgupta, and Peter Henry.
\newblock End-to-end learning of geometry and context for deep stereo
  regression.
\newblock In {\em Computer Vision and Pattern Recognition (CVPR)}, 2017.

\bibitem{luoke2019pmvsnet}
Luo Keyang, Guan Tao, Ju Lili, Huang Haipeng, and Luo Yawei.
\newblock Pmvsnet: Learning patch-wise matching confidence aggregation for
  multi-view stereo.
\newblock In {\em International Conference on Computer Vision (ICCV)}, 2019.

\bibitem{knapitsch2017tanks}
Arno Knapitsch, Jaesik Park, Qian-Yi Zhou, and Vladlen Koltun.
\newblock Tanks and temples: Benchmarking large-scale scene reconstruction.
\newblock In {\em ACM Transactions on Graphics (TOG)}, 2017.

\bibitem{li2018megadepth}
Zhengqi Li and Noah Snavely.
\newblock Megadepth: Learning single-view depth prediction from internet
  photos.
\newblock In {\em Computer Vision and Pattern Recognition (CVPR)}, 2018.

\bibitem{luo2019contextdesc}
Zixin Luo, Tianwei Shen, Lei Zhou, Jiahui Zhang, Yao Yao, Shiwei Li, Tian Fang,
  and Long Quan.
\newblock Contextdesc: Local descriptor augmentation with cross-modality
  context.
\newblock In {\em Computer Vision and Pattern Recognition (CVPR)}, 2019.

\bibitem{luo2018geodesc}
Zixin Luo, Tianwei Shen, Lei Zhou, Siyu Zhu, Runze Zhang, Yao Yao, Tian Fang,
  and Long Quan.
\newblock Geodesc: Learning local descriptors by integrating geometry
  constraints.
\newblock In {\em European Conference on Computer Vision (ECCV)}, 2018.

\bibitem{luo2020aslfeat}
Zixin Luo, Lei Zhou, Xuyang Bai, Hongkai Chen, Jiahui Zhang, Yao Yao, Shiwei
  Li, Tian Fang, and Long Quan.
\newblock Aslfeat: Learning local features of accurate shape and localization.
\newblock In {\em Computer Vision and Pattern Recognition (CVPR)}, 2020.

\bibitem{mayer2016large}
Nikolaus Mayer, Eddy Ilg, Philip Hausser, Philipp Fischer, Daniel Cremers,
  Alexey Dosovitskiy, and Thomas Brox.
\newblock A large dataset to train convolutional networks for disparity,
  optical flow, and scene flow estimation.
\newblock In {\em Computer Vision and Pattern Recognition (CVPR)}, 2016.

\bibitem{merrell2007real}
Paul Merrell, Amir Akbarzadeh, Liang Wang, Philippos Mordohai, Jan-Michael
  Frahm, Ruigang Yang, David Nist{\'e}r, and Marc Pollefeys.
\newblock Real-time visibility-based fusion of depth maps.
\newblock In {\em International Conference on Computer Vision (ICCV)}, 2007.

\bibitem{paschalidou2018raynet}
Despoina Paschalidou, Osman Ulusoy, Carolin Schmitt, Luc Van~Gool, and Andreas
  Geiger.
\newblock Raynet: Learning volumetric 3d reconstruction with ray potentials.
\newblock In {\em Computer Vision and Pattern Recognition (CVPR)}, 2018.

\bibitem{richter2016playing}
Stephan~R Richter, Vibhav Vineet, Stefan Roth, and Vladlen Koltun.
\newblock Playing for data: Ground truth from computer games.
\newblock In {\em European Conference on Computer Vision (ECCV)}, 2016.

\bibitem{ros2016synthia}
German Ros, Laura Sellart, Joanna Materzynska, David Vazquez, and Antonio~M
  Lopez.
\newblock The synthia dataset: A large collection of synthetic images for
  semantic segmentation of urban scenes.
\newblock In {\em Computer Vision and Pattern Recognition}, 2016.

\bibitem{schonberger2016structure}
Johannes~L Schonberger and Jan-Michael Frahm.
\newblock Structure-from-motion revisited.
\newblock In {\em Computer Vision and Pattern Recognition (CVPR)}, 2016.

\bibitem{schonberger2016pixelwise}
Johannes~L Sch{\"o}nberger, Enliang Zheng, Jan-Michael Frahm, and Marc
  Pollefeys.
\newblock Pixelwise view selection for unstructured multi-view stereo.
\newblock In {\em European Conference on Computer Vision (ECCV)}, 2016.

\bibitem{schoeps2017cvpr}
Thomas Sch\"ops, Johannes~L. Sch\"onberger, Silvano Galliani, Torsten Sattler,
  Konrad Schindler, Marc Pollefeys, and Andreas Geiger.
\newblock A multi-view stereo benchmark with high-resolution images and
  multi-camera videos.
\newblock 2017.

\bibitem{seitz2006comparison}
Steven~M Seitz, Brian Curless, James Diebel, Daniel Scharstein, and Richard
  Szeliski.
\newblock A comparison and evaluation of multi-view stereo reconstruction
  algorithms.
\newblock In {\em Computer Vision and Pattern Recognition (CVPR)}, 2006.

\bibitem{shen2018mirror}
Tianwei Shen, Zixin Luo, Lei Zhou, Runze Zhang, Siyu Zhu, Tian Fang, and Long
  Quan.
\newblock Matchable image retrieval by learning from surface reconstruction.
\newblock In {\em Asian Conference on Computer Vision (ACCV}, 2018.

\bibitem{straub2019replica}
Julian Straub, Thomas Whelan, Lingni Ma, Yufan Chen, Erik Wijmans, Simon Green,
  Jakob~J Engel, Raul Mur-Artal, Carl Ren, Shobhit Verma, et~al.
\newblock The replica dataset: A digital replica of indoor spaces.
\newblock {\em arXiv preprint arXiv:1906.05797}, 2019.

\bibitem{strecha2008benchmarking}
Christoph Strecha, Wolfgang von Hansen, Luc Van~Gool, Pascal Fua, and Ulrich
  Thoennessen.
\newblock On benchmarking camera calibration and multi-view stereo for high
  resolution imagery.
\newblock In {\em Computer Vision and Pattern Recognition (CVPR)}, 2008.

\bibitem{tremblay2018training}
Jonathan Tremblay, Aayush Prakash, David Acuna, Mark Brophy, Varun Jampani, Cem
  Anil, Thang To, Eric Cameracci, Shaad Boochoon, and Stan Birchfield.
\newblock Training deep networks with synthetic data: Bridging the reality gap
  by domain randomization.
\newblock In {\em Computer Vision and Pattern Recognition Workshops (CVPRW)},
  2018.

\bibitem{wu2018group}
Yuxin Wu and Kaiming He.
\newblock Group normalization.
\newblock In {\em European Conference on Computer Vision (ECCV)}, 2018.

\bibitem{xue2019mvscrf}
Youze Xue, Jiansheng Chen, Weitao Wan, Yiqing Huang, Cheng Yu, Tianpeng Li, and
  Jiayu Bao.
\newblock Mvscrf: Learning multi-view stereo with conditional random fields.
\newblock In {\em International Conference on Computer Vision (ICCV)}, 2019.

\bibitem{yao2018mvsnet}
Yao Yao, Zixin Luo, Shiwei Li, Tian Fang, and Long Quan.
\newblock Mvsnet: Depth inference for unstructured multi-view stereo.
\newblock In {\em European Conference on Computer Vision (ECCV)}, 2018.

\bibitem{yao2019recurrent}
Yao Yao, Zixin Luo, Shiwei Li, Tianwei Shen, Tian Fang, and Long Quan.
\newblock Recurrent mvsnet for high-resolution multi-view stereo depth
  inference.
\newblock In {\em Computer Vision and Pattern Recognition (CVPR)}, 2019.

\bibitem{zhang2018unreal}
Yi Zhang, Weichao Qiu, Qi Chen, Xiaolin Hu, and Alan Yuille.
\newblock Unrealstereo: Controlling hazardous factors to analyze stereo vision.
\newblock In {\em International Conference on 3D Vision)}, 2018.

\end{thebibliography}
}

\newpage

\twocolumn[
\begin{center}
\textbf{\LARGE Supplemental Materials}
\vspace{5mm}
\end{center}
{
	\renewcommand\twocolumn[1][]{#1}%
	\begin{center}
	\centering
	\includegraphics[width=1\linewidth]{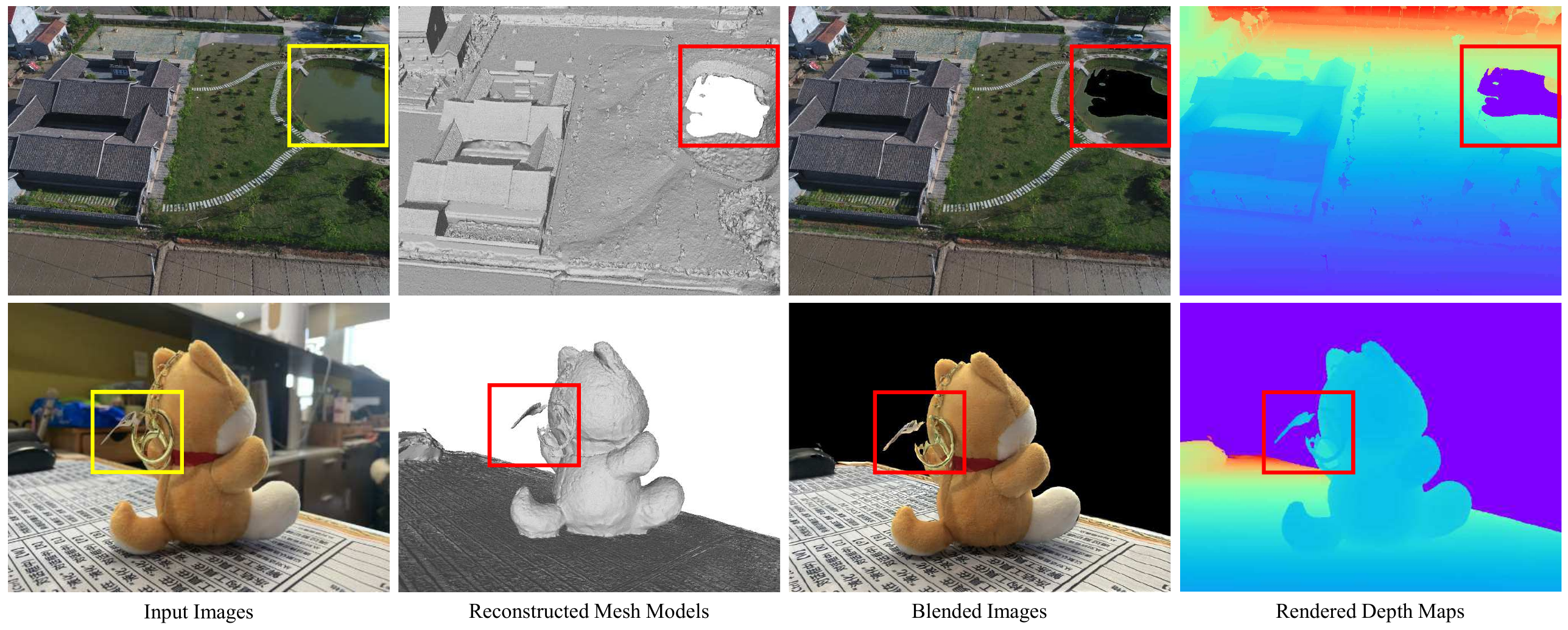}
	\captionof{figure}{Reconstruction defects in BlendedMVS datasets.}
  	\label{fig:defect}
	\end{center}%
}
]
\setcounter{section}{0}
\setcounter{equation}{0}
\setcounter{figure}{0}
\setcounter{table}{0}
\setcounter{page}{1}
\makeatletter

\section{Imperfect Reconstruction}

We visualize two imperfect mesh reconstructions during the data generation process in Fig.~\ref{fig:defect}. The two yellow regions are a reflective water pond and a thin key ring respectively. We failed to reconstruct these two regions in the mesh models, and the resulting blended images and rendered depth maps are also incomplete. However, as discussed in Sec.~5.3, such reconstruction defects won't affect the network training, because we will not use the original images as training inputs, and the blended images we used could be consistent aligned to rendered depth maps.

\section{Input v.s. Blended v.s. Rendered Images}
\begin{figure*}[!h]
  \centering
  \includegraphics[width=0.92\linewidth]{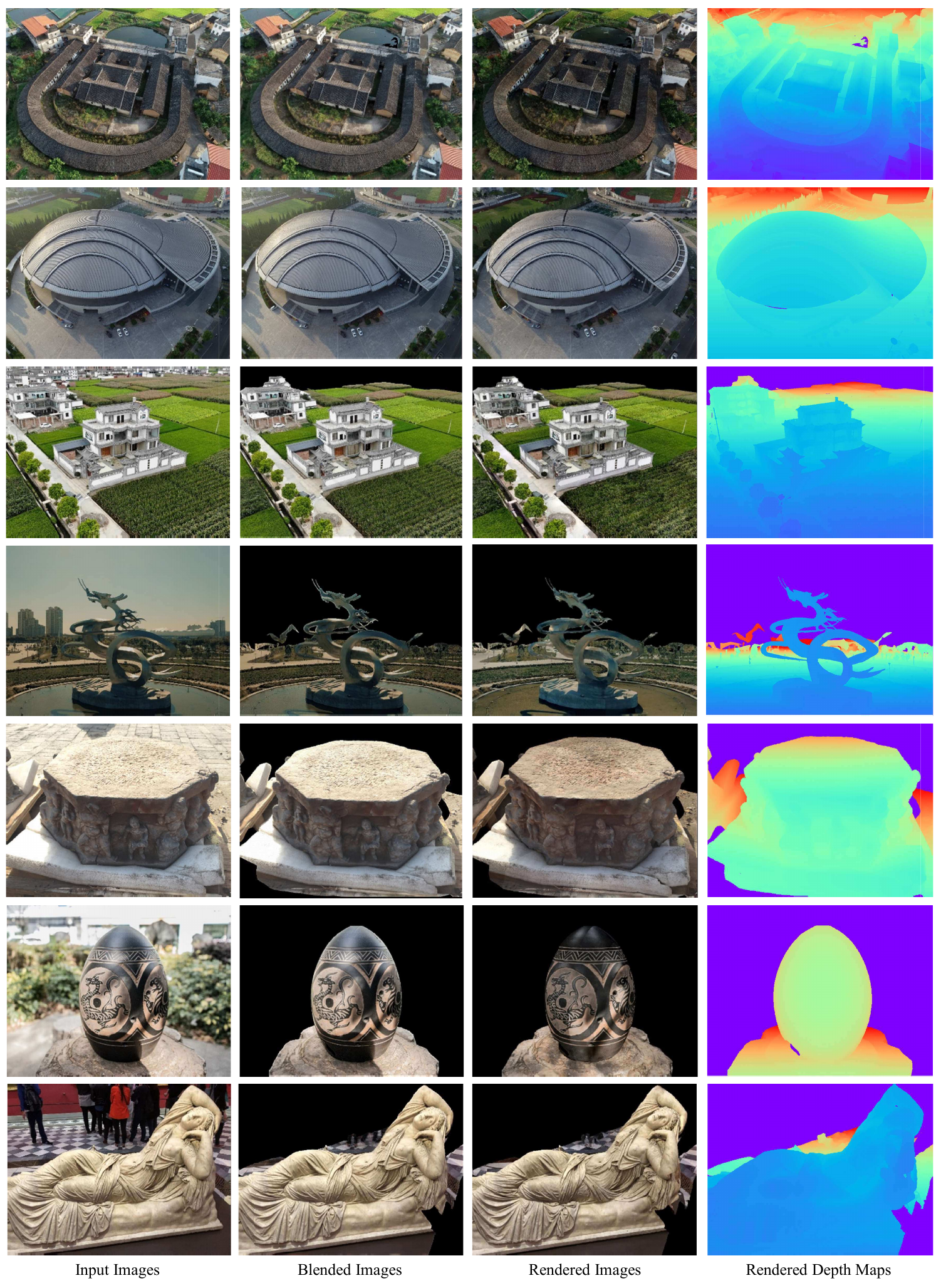}
  \vspace{-2mm}
  \caption{Comparison among input images, blended images and rendered images.}
  \label{fig:images}
\end{figure*}

In this section, we further illustrate the difference between 1) input images, 2) blended images, and 3) rendered images. The corresponding rendered depth maps are also jointly visualized in Fig.~\ref{fig:images}. The blended image has similar lightings to the input image, and also inherits detailed visual cues from the rendered image.

\section{Scenes}
\begin{figure*}
  \centering
  \includegraphics[width=0.92\linewidth]{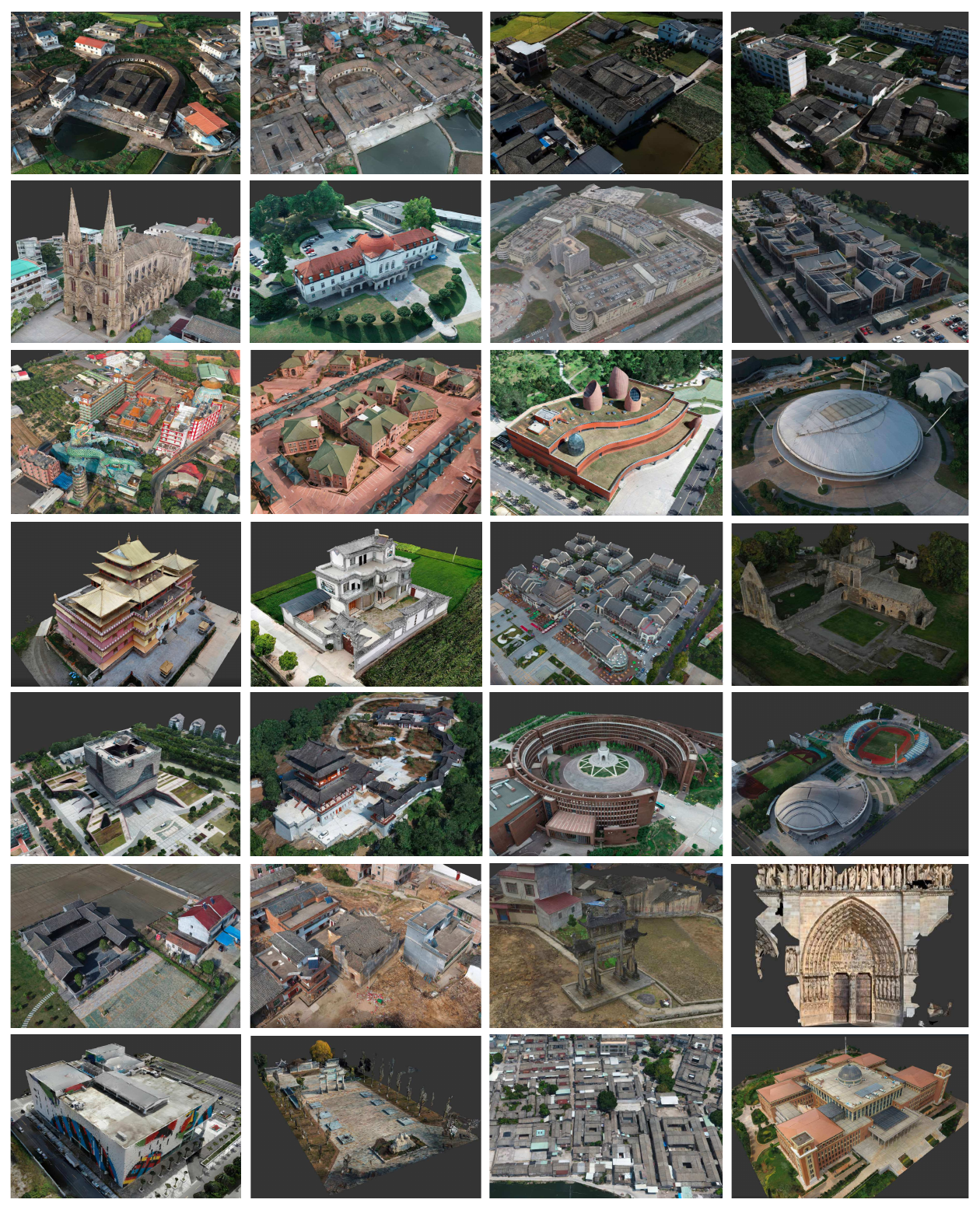}
  \vspace{-2mm}
  \caption{Large-scale outdoor scenes in BlendedMVS dataset.}
  \label{fig:large}
\end{figure*}

\begin{figure*}
  \centering
  \includegraphics[width=0.92\linewidth]{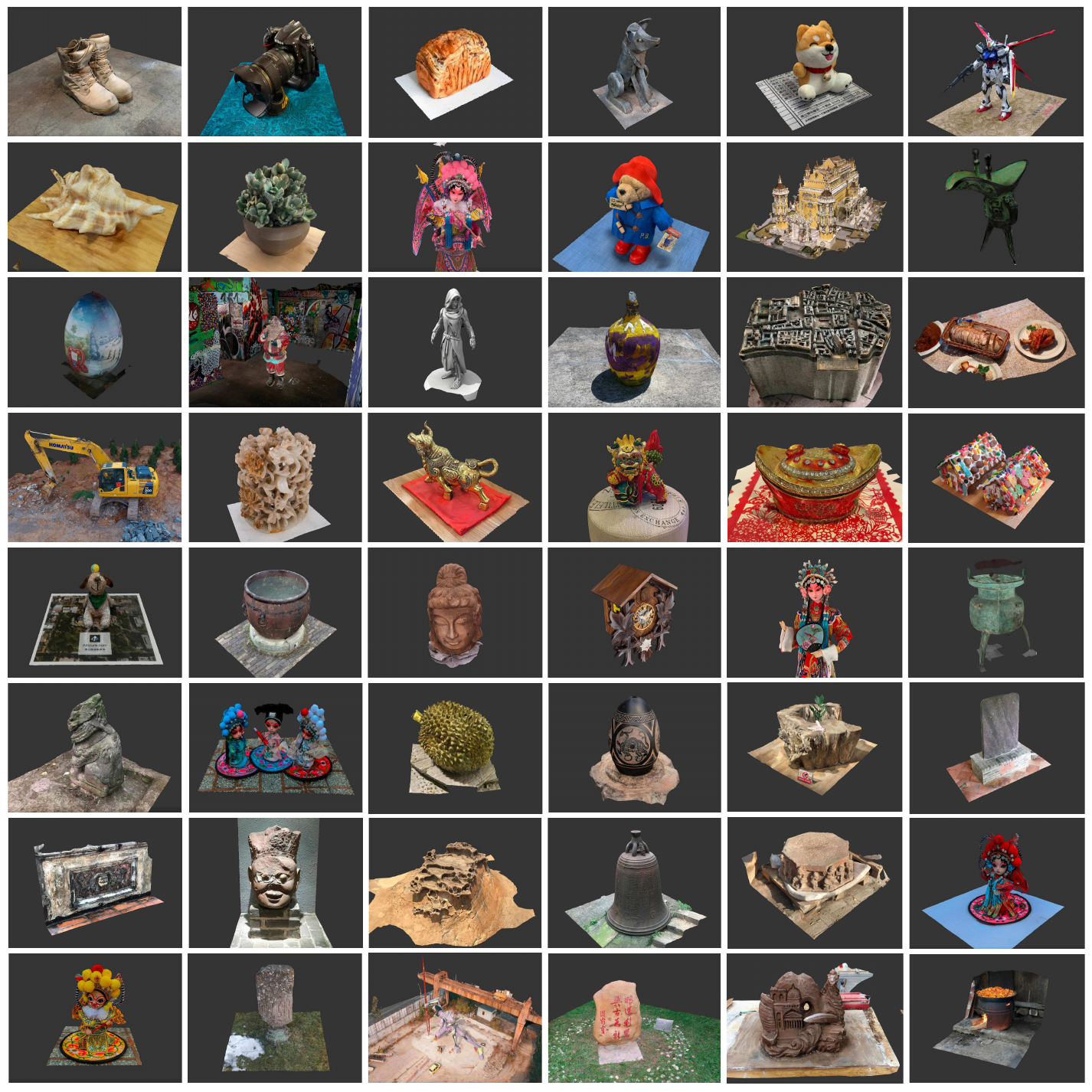}
  \vspace{-2mm}
  \caption{Small-scale objects in BlendedMVS dataset.}
  \label{fig:small}
\end{figure*}

\begin{figure*}
  \centering
  \includegraphics[width=0.92\linewidth]{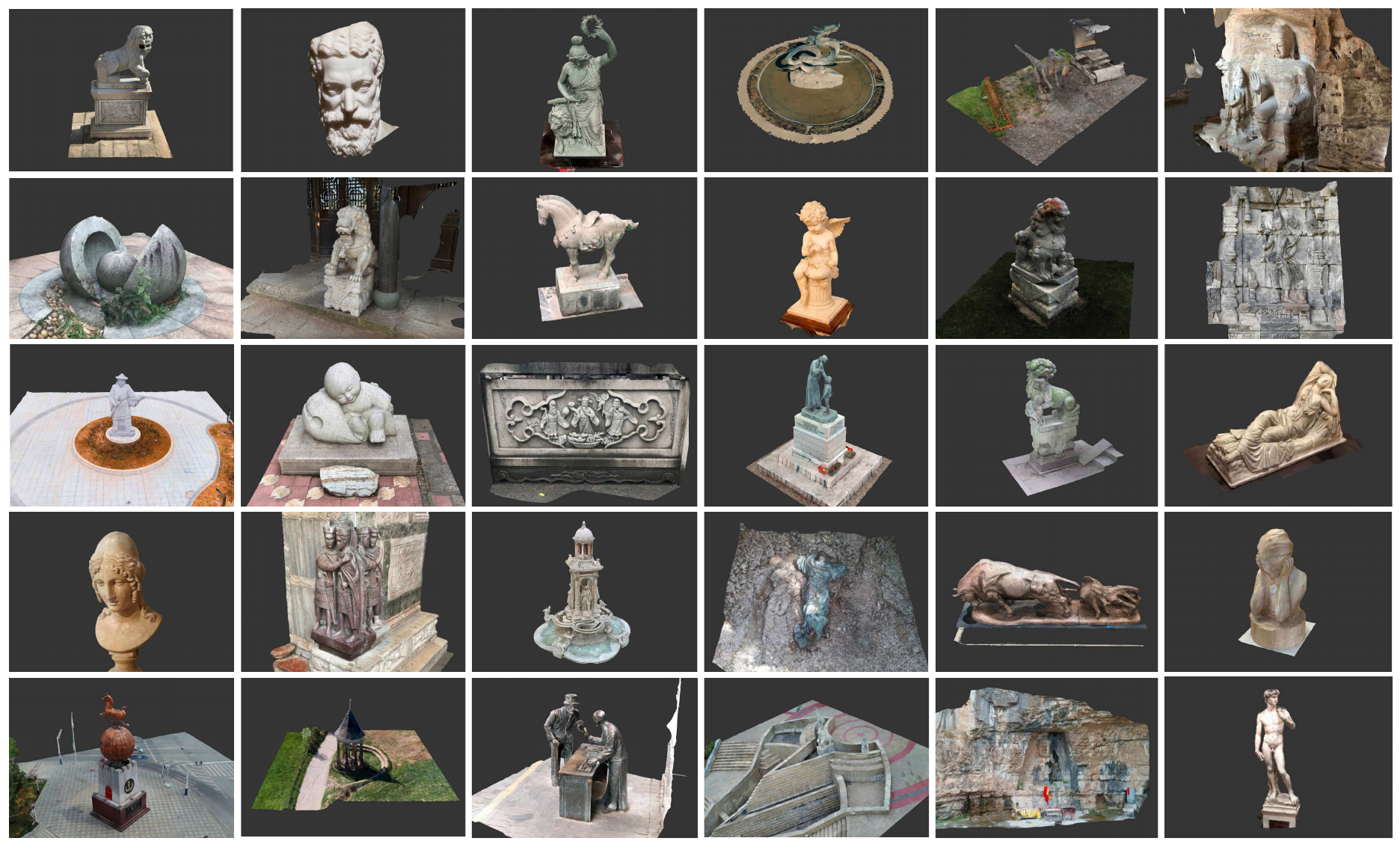}
  \vspace{-2mm}
  \caption{High-quality sculptures in BlendedMVS datasets.}
  \label{fig:sculpture}
\end{figure*}

In this section, we list all textured models used in BlendedMVS dataset. We manually crop the boundary areas of some scenes for better visualization. Models in our dataset can be roughly categorized into 1) large-scale scenes in Fig.~\ref{fig:large}, 2) small-scale objects in Fig.~\ref{fig:small}, and 3) high-quality sculptures in Fig.~\ref{fig:sculpture}. Also, the 7 validation scenes are shown in Fig.~\ref{fig:validation}.

\begin{figure*}
  \centering
  \includegraphics[width=0.92\linewidth]{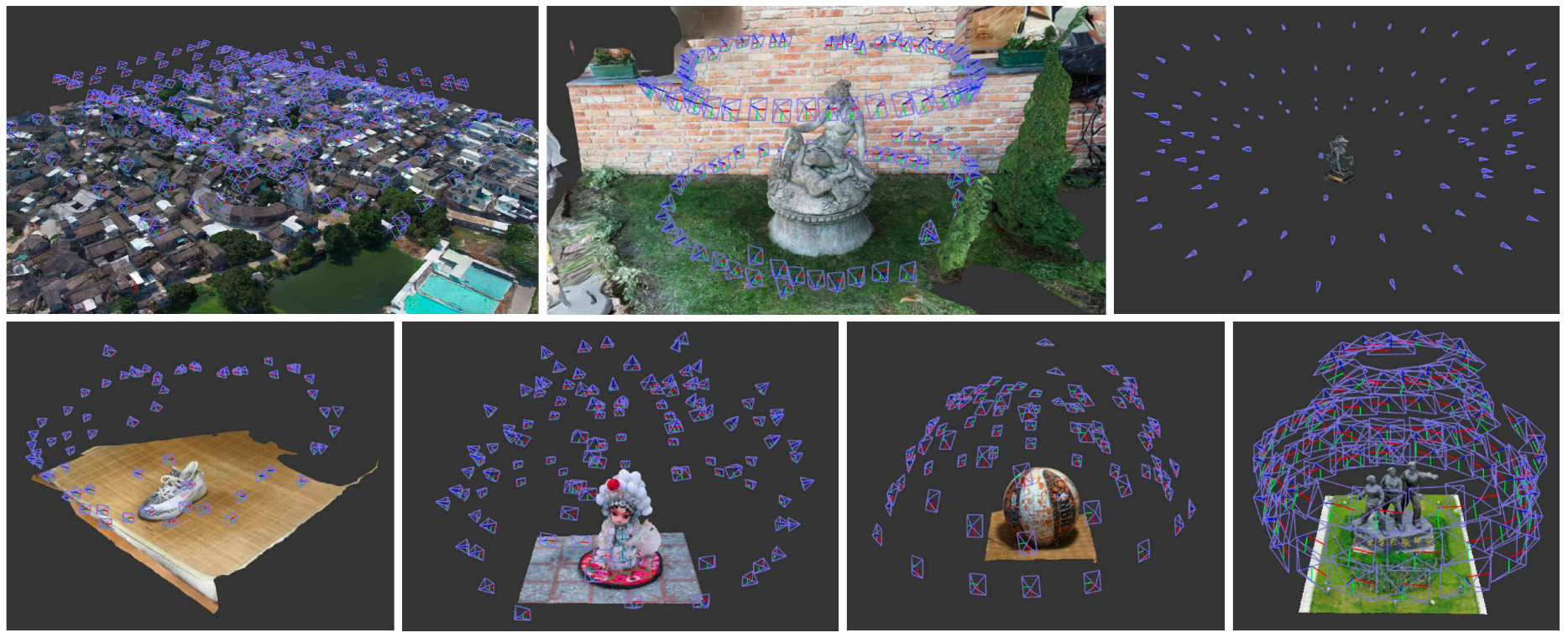}
  \vspace{-2mm}
  \caption{Validation scenes with camera trajectories.}
  \label{fig:validation}
\end{figure*}

\end{document}